\documentclass[conference]{IEEEtran}
\IEEEoverridecommandlockouts
\usepackage{cite}
\usepackage{amsmath,amssymb,amsfonts}
\usepackage{algorithmic}
\usepackage{graphicx}
\usepackage{textcomp}
\usepackage{xcolor}
\usepackage{comment}
\def\BibTeX{{\rm B\kern-.05em{\sc i\kern-.025em b}\kern-.08em
    T\kern-.1667em\lower.7ex\hbox{E}\kern-.125emX}}

\begin{document}

\title{Exploration of the generative capabilities of Boltzmann machines applied to social systems under the majority rule\\

\thanks{The authors thank ANID FONDECYT 1250386, ANID FONDECYT 1230315, ANID-MILENIO-NCN2024\_103, ANID-MILENIO-NCN2024\_047, and Centro de Modelamiento Matemático (CMM) FB210005, BASAL funds for centers of excellence from ANID-Chile.}

}

\author{\IEEEauthorblockN{Mauricio A. Valle, Gonzalo A. Ruz}}

\author{\IEEEauthorblockN{1\textsuperscript{st} Mauricio A. Valle}
\IEEEauthorblockA{\textit{Faculty of Engineering and Sciences} \\
\textit{Universidad Adolfo Ib\'{a}\~{n}ez}\\
\textit{Millennium Nucleus for Social Data Science (SODAS)}\\
Santiago, Chile \\
mauricio.valle.b@uai.cl}
\and
\IEEEauthorblockN{2\textsuperscript{nd} Gonzalo A. Ruz}
\IEEEauthorblockA{\textit{Faculty of Engineering and Sciences} \\
\textit{Universidad Adolfo Ib\'{a}\~{n}ez}\\
\textit{Millennium Nucleus for Social Data Science (SODAS)}\\
\textit{Millennium Nucleus in Data Science for}\\
\textit{Plant Resilience (PhytoLearning)}\\
Santiago, Chile \\
gonzalo.ruz@uai.cl}
}

\maketitle

\begin{abstract}
We study the generative capabilities of Boltzmann machines to recover systems governed by the majority rule under critical conditions. To this end, we train deep belief networks (DBNs) with different configurations, where the first layer can use Gaussian visible units with more than two states (i.e., non-binary units). We then allow the DBN to ``dream'' samples conditioned on visible units that we keep fixed, and we measure the deviation of this dreamed system from the real one. We also corroborate, using a discrete thermometer based on a convolutional network, that the reconstructions remain in a critical state. Across several training sessions with different architectures, we show that, despite the complexity of the problem, the DBN can recover samples that remain critical even under input noise, with a gradual degradation of physical observables relative to the original sample.
\end{abstract}


\section{Introduction}

We are constantly subjected to opinions that change as new information arrives and according to the opinions of other people with whom we interact. At the collective level, the dynamics of opinions of a group of people who interact are subject to mechanisms that give rise to complex and difficult-to-predict patterns. From applied mathematics and physics, including theories from sociology and economics, algorithms have been developed that attempt to reproduce social phenomena such as consensus, polarization, and fragmentation of opinions \cite{peralta}.

In financial markets, for instance, agents driven by both rational and emotional behaviors are difficult to predict and simulate. Agents who interact with their peers tend to adopt herd behavior, following what the majority does because they feel more comfortable when their decisions are supported by other people. This produces cycles of self-reinforcement that lead to over-reactions to good or bad news, translating into abrupt changes in decisions to buy and sell financial assets \cite{vilela,hohnisch,crescimanna}. This manifests itself in regime shifts or phase transitions in which a system that was in a state of relative calm begins to change its state to one of great turmoil and high variability. Phase transitions represent critical tipping points in social systems where collective behavior undergoes fundamental shifts.  Understanding these transitions, the moments when systems shift from disorder to order, or from consensus to polarization, is crucial for predicting and potentially mitigating extreme collective behaviors such as market crashes, rapid opinion polarization in social networks, or the sudden emergence of social movements \cite{castellano,tsarev}. Phase transitions in social systems are not merely academic curiosities; they represent moments of maximum uncertainty and maximum potential for change, making them both the most challenging and the most important phenomena to understand \cite{mansouri}.

The intersection of statistical physics and machine learning has opened promising new avenues for understanding complex systems. In particular, Restricted Boltzmann Machines (RBMs) are energy-based probabilistic neural networks that learn a distribution over binary (or real-valued) variables via latent hidden units, and Deep Belief Networks (DBNs) are multilayer generative models formed by stacking RBMs. Deep learning models based on energy-based frameworks offer a natural language for describing systems governed by collective interactions. Boltzmann machines can capture latent representations of spin configurations at different temperatures and identify phase transitions in many-body systems \cite{funai,funai2,wang,hu,gu,valle}. DBNs and Gaussian-Bernoulli Restricted Boltzmann Machines (GBRBMs) are particularly well-suited for analyzing opinion dynamics models for several compelling reasons. First, these models are inherently probabilistic and energy-based, making them conceptually aligned with the statistical mechanics framework that underlies opinion dynamics. The Boltzmann distribution that governs RBMs mirrors the equilibrium distributions found in statistical physics models. Second, unlike discriminative models that simply learn input-output mappings, generative models like DBNs learn the underlying probability distribution of the data itself. This means they can capture the intrinsic structure of different phases, subcritical, critical, and supercritical, without being explicitly told about phase boundaries. Third, the hierarchical architecture of DBNs naturally corresponds to the multi-scale nature of phase transitions, where local interactions give rise to global phenomena. The visible layer can represent microscopic agent states, while hidden layers progressively capture increasingly abstract features such as cluster formation, correlation lengths, and system-wide order parameters.

This work addresses a critical gap in our understanding of how unsupervised learning models can extract meaningful representations from highly unstructured social systems. Unlike well-studied image datasets such as MNIST, where spatial structure provides clear regularities, opinion dynamics systems exhibit heterogeneous, irregular patterns, particularly near phase transitions where the system balances between order and disorder \cite{balankin,lima,zubillaga,vilela2}. Our investigation has three key contributions. First, we show that DBNs can successfully learn compressed representations of the three-state majority-vote model (MV3) across different phases, achieving low reconstruction errors despite the configurational complexity. Second, we develop and validate a convolutional ``thermometer'' network capable of classifying system states as subcritical, critical, or supercritical, effectively learning to recognize phase signatures without explicit knowledge of order parameters. Third, and perhaps most importantly, we explore the generative capabilities of DBNs through systematic imputation experiments, revealing how these models reconstruct missing information and maintain criticality even under substantial noise. This capability has profound implications: if machine learning models can reliably ``dream'' realistic samples of social systems at criticality, they can potentially serve as powerful tools for scenario generation, early warning systems for phase transitions, and hypothesis testing in computational social science. From a practical point of view, we envision a machine learning system capable of reconstructing agent opinions from incomplete information, with immediate applications in real-world scenarios where complete data is rarely available. In social media analytics, for instance, we typically observe only a fraction of users' opinions (those who actively post), while the "silent majority" remains unobserved. A trained DBN could impute these missing opinions, providing a more complete picture of the opinion landscape and identifying brewing consensus or polarization before it becomes visible in public discourse.


In this paper, we present a comprehensive investigation of DBN capabilities for the MV3 model. Although our simulations necessarily simplify many aspects of real complex social systems by using regular lattices rather than realistic network topologies, our focus is on probing the generative and reconstructive capabilities of Boltzmann machines when applied to majority-rule systems with three possible states. We show that, despite the complexity of the problem, DBNs can recover samples that remain in critical states under imposed noise, with a gradual degradation of physical observables relative to the original samples. This establishes a foundation for future extensions to more realistic network structures and opens new possibilities for machine learning--driven analysis of social phase transitions.



This paper is organized as follows. The proposed approach is described and explained in Section II, as well the results of our simulations. Finally, the paper concludes in Section III.

\section{Proposed Method}
\begin{figure}[t]
\centerline{\includegraphics[scale=0.5]{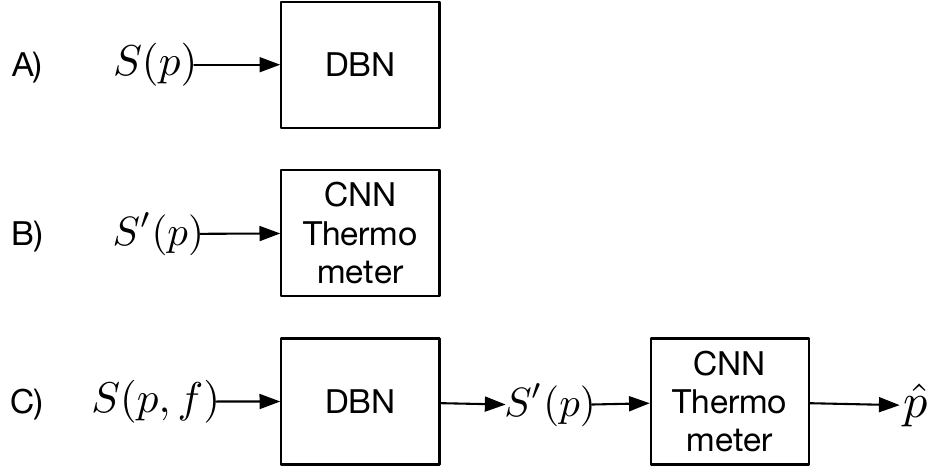}}
\caption{Methodology developed to evaluate the generative capabilities of the DBN.}
\label{fig1}
\end{figure}

To investigate the unsupervised learning capabilities of DBN, we chose to simulate systems consisting of agents with discrete opinions, governed by the majority rule. These systems are highly unstructured and heterogeneous, unlike other well-known cases such as handwritten digits or fashion products in the MNIST database. After creating a sufficient number of samples, we can use them to A) train different DBN architectures, B) train a convolutional network as a thermometer to determine the condition of the system (for example, in a condition close to criticality or supercriticality), and finally C) use samples to conduct experiments with the DBN to generate synthetic samples of systems in a critical condition with a certain fraction $f$ of noise.  This allows us to test the level of corruption a system can have before the DBN fails to recover it properly as a sample in critical condition. It is important to mention that simulated system samples are not shared for the various activities indicated above.
The methodology used in this study follows the activities indicated in Figure \ref{fig1}. 

\subsection{MV3 system and Generation of dataset}
\begin{figure}[t]
\centerline{\includegraphics[scale=0.55]{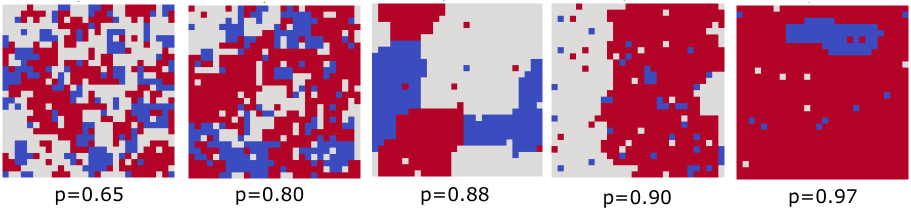}}
\caption{Some examples of systems $S(p)$ of $L=28^2=784$ for different values of the parameter $p$, from more to less noise.}
\label{fig2}
\end{figure}

\begin{figure*}[t]
\centerline{\includegraphics[scale=0.38]{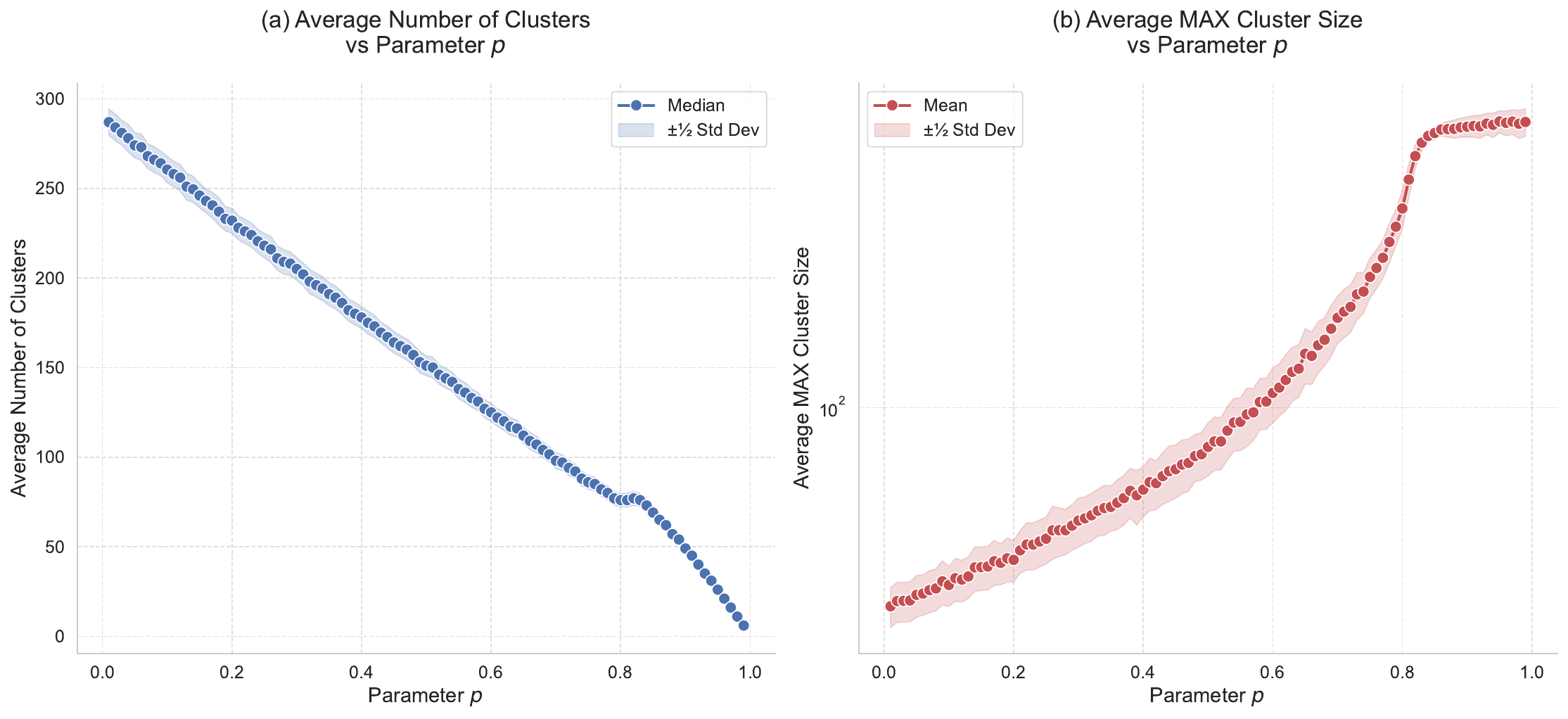}}
\caption{Cluster behavior in the MV3 system of $28 \times 28$. Left: Average number of clusters found in the system for any opinion type. Right: Maximum cluster size found in the system independent of opinion. For each value of $p$, a simulation was performed 30 times to calculate a confidence band of $\pm 1 sd$.}
\label{clusters}
\end{figure*}

We used the majority vote model with a version of three states (MV3 model) to simulate the state of opinions of agents located on a square lattice with regular boundary conditions. On this lattice, there are $L=N^2$ agents that can be in one of three states or opinions represented by $s_i \in \{ 0, 0.5, 1 \}$. Each agent has four neighbors. The state of the system is represented by $S(p)= (s_1, s_2, ..., s_N)$. The energy of the system is 
\begin{equation}
    H(S)= \sum_{(i,j)} \delta(s_i, s_j)
\end{equation}
\noindent where $\delta$ is the Kronecker delta, whose value is 1 if $s_i=s_j$ or 0 otherwise. To simulate the system, at each time, the states of the agents are updated simultaneously according to the following local rule: a) If the agent's neighbors have a majority opinion, then the agent's opinion changes to that of the majority with probability $p=1-q$, otherwise it disagrees with the majority and continues with the same opinion with probability $q=1-p$, b) If there is no majority, then the agent assumes either of the two states with equal probability. We perform Monte Carlo simulations on lattices with size $L=28^2=784$ for a given value of $p$. A complete update of the 784 states of the agents corresponds to a Monte Carlo step (MCS) equivalent to one unit of time in our simulations. Each simulation started with a configuration generated at random. We let the simulation run freely for at least 100,000 MCS to allow the system to reach a steady state before taking a sample of the system $S(p)$. We repeated this process 79,000 times for different values of $p$. The parameter $q= 1-p$ can be considered as the noise level of the system. For high values of $q$ (low probability of agreement), the system tends to be disordered, while for low values of $q$ (high probability of agreement), the system tends to become ordered, adopting a majority opinion among the agents equivalent to a consensus. The observed critical noise value for this system is $q_c=0.106$ (or $p=0.894$) \cite{lima}, at which a phase transition is observed, characterized by the coexistence of clusters of two or three opinion states simultaneously, giving rise to a variety of polarization possibilities. In our case, due to finite-size effects of the grid we use, the transition point is slightly shifted to $q_c=0.151$ (or $p=0.849$), resulting in a more diffuse or less abrupt transition compared to much larger grid sizes. Figure \ref{fig2} shows an example of $S(p)$ systems for different values of the parameter $p$. This phenomenon can also be visualized in Figure \ref{clusters}(left): as the probability of agreement $p$ decreases, the average number of opinion clusters increases, indicating the fragmentation of opinion groups into smaller units. The effect is even clearer for the maximum cluster size in Figure \ref{clusters}(right). For $p>p_c$, consensus is practically reached (the maximum cluster size approaches $28 \times 28 = 784$). However, near $p_c$ the cluster sizes become unstable, producing a rapid decrease in the maximum size for $p<p_c$.

\subsection{Training of DBNs}
Unlike classic opinion systems in which agents can only have two opinions (e.g., for or against), we consider the possibility that agents can have three possible states, similar to what occurs, for example, with investors, whose opinion regarding the market can be bullish, bearish, or neutral. Standard binary RBMs cannot adequately represent this tri-state structure, so we decided to train a variant of this machine in which the visible units can be Gaussian, and we left the hidden units binary called Gaussian-Bernoulli RBM (GBRBM). The~energy function is given by

\begin{equation}
\begin{aligned}
    E(v,h) &= -\sum_{i,j} w_{ij} \frac{h_j}{s_j} \frac{v_i}{\sigma_i} - \sum_{i}^{m} \frac{(v_i-b_i)^2}{2 \sigma_i} - \\
      &\sum_{j}^{n} \frac{(h_j  - c_j)^2}{2s_j}
\end{aligned}
\end{equation}

\noindent where $v_i$ and $h_j$ are real values for the visible and binary hidden units, respectively. The parameters $b_i$ and $c_j$ are biases to the visible and hidden units, respectively, connected through the weights $w_{ij}$. Note that the visible units $v_i$ are equivalent to the states of the system for $s_i$ for any agent in the grid defined previously.

The~hyperparameters $\sigma_i$ and $s_j$ are the standard deviations of the visible and hidden units, respectively. The~conditional probabilities follow these equations \cite{liao}:

\begin{equation}\label{eqpvh2}
    p(\mathbf{v}|\mathbf{h}) = N\Bigl( \mathbf{v}|\mathbf{b} + \sum_{j} h_j w_{ij}, \Sigma^2 \Bigr),
\end{equation}
\begin{equation}\label{eqphv2}
    p(\mathbf{h}|\mathbf{v}) = N\Bigl( \mathbf{h}|\mathbf{c} + \sum_{i} v_i w_{ij}, S^2  \Bigr)
\end{equation}

\noindent where $N(\mu, \Sigma)$ represents the multivariate normal distribution with mean $\mu$ and covariance $\Sigma^2$, respectively, while $S^2$ is the covariance matrix of hidden units. 
We leave both covariance matrices fixed \cite{karakida}. For the visible units, we use the empirical covariance matrix of the training data, while for the hidden units, we set $S = sI_m$, assuming statistical independence between the hidden units. We found that setting $s = 0.5$ yielded the most stable convergence.

After training the GBRBM, we use the hidden units of this network as inputs to train a second layer in a traditional RBM, i.e., the hidden units of the first layer serve as the visible units to train the next RBM, or a Bernoulli-Bernoulli RBM (BBRBM). In this case, the conditional probabilities for updating the visible and hidden units follow the regular expression \cite{fisher}. Thus, the DBN consists of several RBMs trained using a greedy, layer-wise, unsupervised learning algorithm.
To estimate the parameter set for each RBM, we use the contrastive divergence (CD) learning approach, which has been the standard for RBM training \cite{hinton02}.

We trained different DBN architectures to compare their capabilities in reconstructing the original systems. Table \ref{tab_arch} shows the representative alternatives from our experiments along with the hyperparameters used. In all cases, the batch size used was 128, the number of Gibbs samplings was $k=1$, the learning rate decay rate was 0.98, the weight regularizer rate is set to 0.01, the decay steps were 20, and finally, the hyperparameter value was $\sigma=0.001$.

\begin{table}[t]
\caption{Reconstruction errors measured on test datasets for different DBN architectures.}
\begin{center}
\begin{tabular}{llllll}
\textbf{Model} & \textbf{Strategy} & \textbf{\begin{tabular}[c]{@{}l@{}}$nH_1$\\ $nH_2$\\ $nH_3$\end{tabular}} & \textbf{epochs} & \textbf{lr} & \textbf{\begin{tabular}[c]{@{}l@{}}recons.\\ error\end{tabular}} \\
\hline
DBN1 & Expand-Contract & \begin{tabular}[c]{@{}l@{}}2704\\ 144\\ 64\end{tabular} & \begin{tabular}[c]{@{}l@{}}400\\ 460\\ 400\end{tabular} & \begin{tabular}[c]{@{}l@{}}0.0001\\ 0.001\\ 0.002\end{tabular} & 0.075 \\
\hline
DBN2 & Expand-Contract & \begin{tabular}[c]{@{}l@{}}2704\\ 144\\ 25\end{tabular} & \begin{tabular}[c]{@{}l@{}}400\\ 460\\ 100\end{tabular} & \begin{tabular}[c]{@{}l@{}}0.0001\\ 0.001\\ 0.001\end{tabular} & 0.080 \\
\hline
DBN3 & Expand-Contract & \begin{tabular}[c]{@{}l@{}}4096\\ 255\\ 144\end{tabular} & \begin{tabular}[c]{@{}l@{}}400\\ 460\\ 100\end{tabular} & \begin{tabular}[c]{@{}l@{}}0.0001\\ 0.001\\ 0.005\end{tabular} & 0.065 \\
\hline
DBN4 & Expand-Contract & \begin{tabular}[c]{@{}l@{}}4096\\ 255\\ 64\end{tabular} & \begin{tabular}[c]{@{}l@{}}400\\ 460\\ 100\end{tabular} & \begin{tabular}[c]{@{}l@{}}0.0001\\ 0.001\\ 0.005\end{tabular} & 0.066 \\
\hline
DBN5 & Compressive & \begin{tabular}[c]{@{}l@{}}784\\ 255\\ 144\end{tabular} & \begin{tabular}[c]{@{}l@{}}400\\ 620\\ 100\end{tabular} & \begin{tabular}[c]{@{}l@{}}0.0001\\ 0.001\\ 0.005\end{tabular} & 0.108 \\
\hline
DBN6 & Expansive & \begin{tabular}[c]{@{}l@{}}784\\ 2401\\ -\end{tabular} & \begin{tabular}[c]{@{}l@{}}400\\ 460\\ -\end{tabular} & \begin{tabular}[c]{@{}l@{}}0.0001\\ 0.001\\ - \end{tabular} & 0.109 \\
\hline
DBN7 & Fully Compress & \begin{tabular}[c]{@{}l@{}}256\\ 64\\ 144\end{tabular} & \begin{tabular}[c]{@{}l@{}}210\\ 460\\ 100\end{tabular} & \begin{tabular}[c]{@{}l@{}}0.0001\\ 0.001\\ 0.005\end{tabular} & 0.132 \\
\hline
\end{tabular}
\label{tab_arch}
\end{center}
\end{table}

\begin{figure}[t]
\centerline{\includegraphics[scale=0.48]{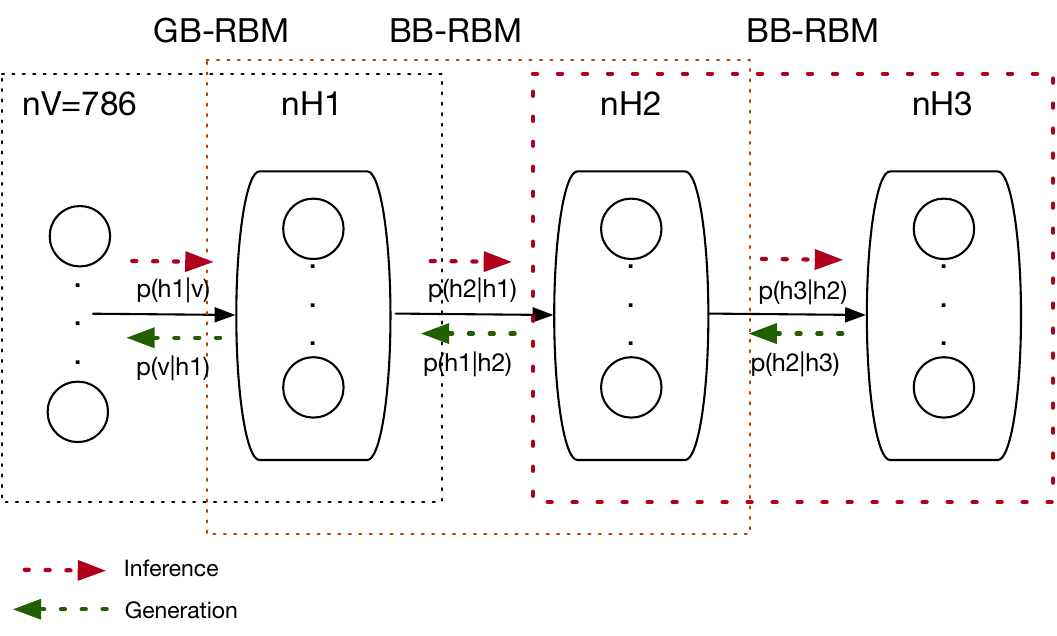}}
\caption{Using the DBN to ``dream'' systems $S'(p)$. Values of $nH1$, $nH2$ and $nH3$ represent the number of units in each stacked layer. In this case, the DBN has three layers, consisting of one GBRBM and then two BBRBMs.}
\label{fig_dbn}
\end{figure}

Next, we take random samples from systems $S(p)$ not used for training, present them to the first input layer of the DBN (i.e., the GBRBM), and let the system generate synthetic samples from them. Figure \ref{fig_dbn} attempts to schematize the process. Table \ref{tab_arch} also shows the reconstruction error achieved by repeating this process for each architecture. The reconstruction error is defined as:
\begin{equation}\label{eq_error}
    \sqrt{ \frac{1}{L} \sum (s_i - \hat{s}_i)^2}
\end{equation}
\noindent where $\hat{s}_i$ is the reconstructed value of agent $i$. As a reference, the worst error that could be expected would be one in which all agents were in consensus in state 0, and the DBN reconstructed a system in consensus with an opposing opinion in state 1. In this case, the error is 1. The second-worst case would be one in which the reconstruction fails for all agents across adjacent states, that is, confusing a 0 with a 0.5 or a 0.5 with a 1, or vice versa. In this case, the error would be 0.5.

\begin{figure}[t]
\centerline{\includegraphics[scale=0.6]{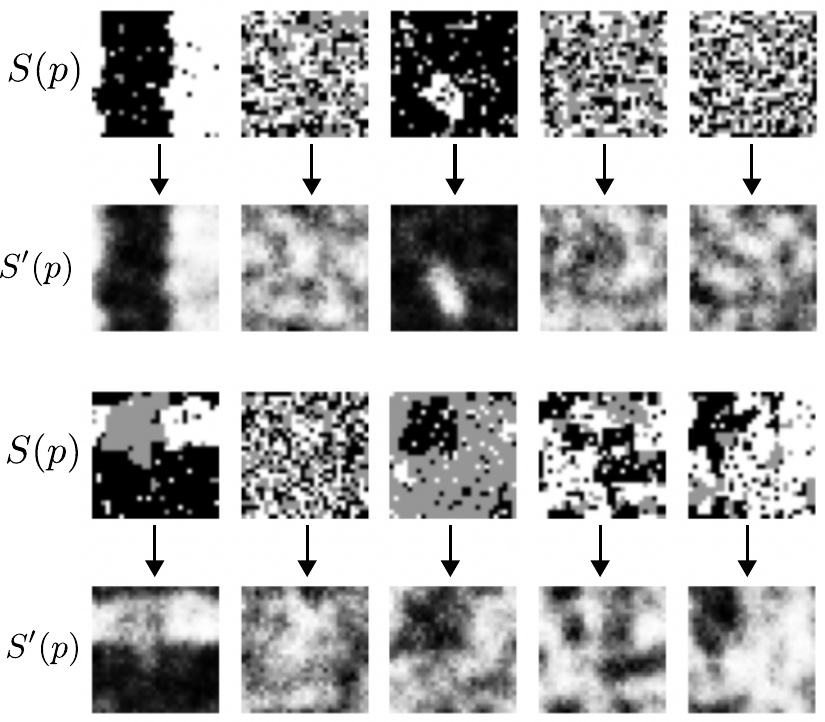}}
\caption{Some examples of reconstruction $S'(p)$ systems given the original $S(p)$ to the DBN. The DBN used here is with $nH1=2704$, $nH2=144$ and $nH3=64$}
\label{fig_recons}
\end{figure}

We see that, despite the enormous number of possible configurations for this type of system, the DBN with a layer of Gaussian units at the beginning can reconstruct the original states acceptably. See, for example, Figure \ref{fig_recons}. It provides a visual comparison of how the DBN-reconstructed samples look compared to the originals.
As we increase the complexity of the DBN by adding more hidden units, the machine can recognize more pattern configurations. In these cases, we use an expansive strategy in which we allow the patterns of the original system to be encoded in a latent space of greater dimension than the original. In summary, Table \ref{tab_arch} reveals three findings: (1) expansive architectures ($nH_1 > 784$) achieve lower reconstruction errors (0.065–0.075), albeit at a higher computational cost; (2) fully compressive architectures ($nH_1 < 784$) sacrifice reconstruction fidelity (0.108–0.132) but exhibit surprising robustness in criticality preservation (Figure \ref{fig_imput_256_144_64}); and (3) expansive-contractive designs balance both objectives. Other contractive architectures (where $nH1<784$), not shown here, produce higher reconstruction errors. Therefore, for sample reconstruction and reproduction, and not necessarily for information compression, an expansive strategy is more useful, though computationally more costly.

\subsection{Training of the discrete thermometer}\label{Section.C}
The thermometer indicates whether the synthetic sample is in a subcritical ($p<0.75$), critical ($0.75 \leq p \leq 0.85$), or supercritical ($p>0.85$) state. This makes it a multi-classification problem. The thermometer is trained on the basis of synthetic samples $S'(p)$ generated by the GBRBM fed by a total of $24,000$ simulated systems exclusively for this task and balanced for each of the three classes. The thermometer architecture is based on three convolutional layers of $20$, $25$, and $30$ channels with $3\times3$ kernels, followed by $2\times2$ max-pooling layers, then two dense layers of $512$ and $128$ units with $0.3$ dropout and RELU activation functions, and an output with a Softmax function. 

\begin{figure}[t]
\centerline{\includegraphics[scale=0.35]{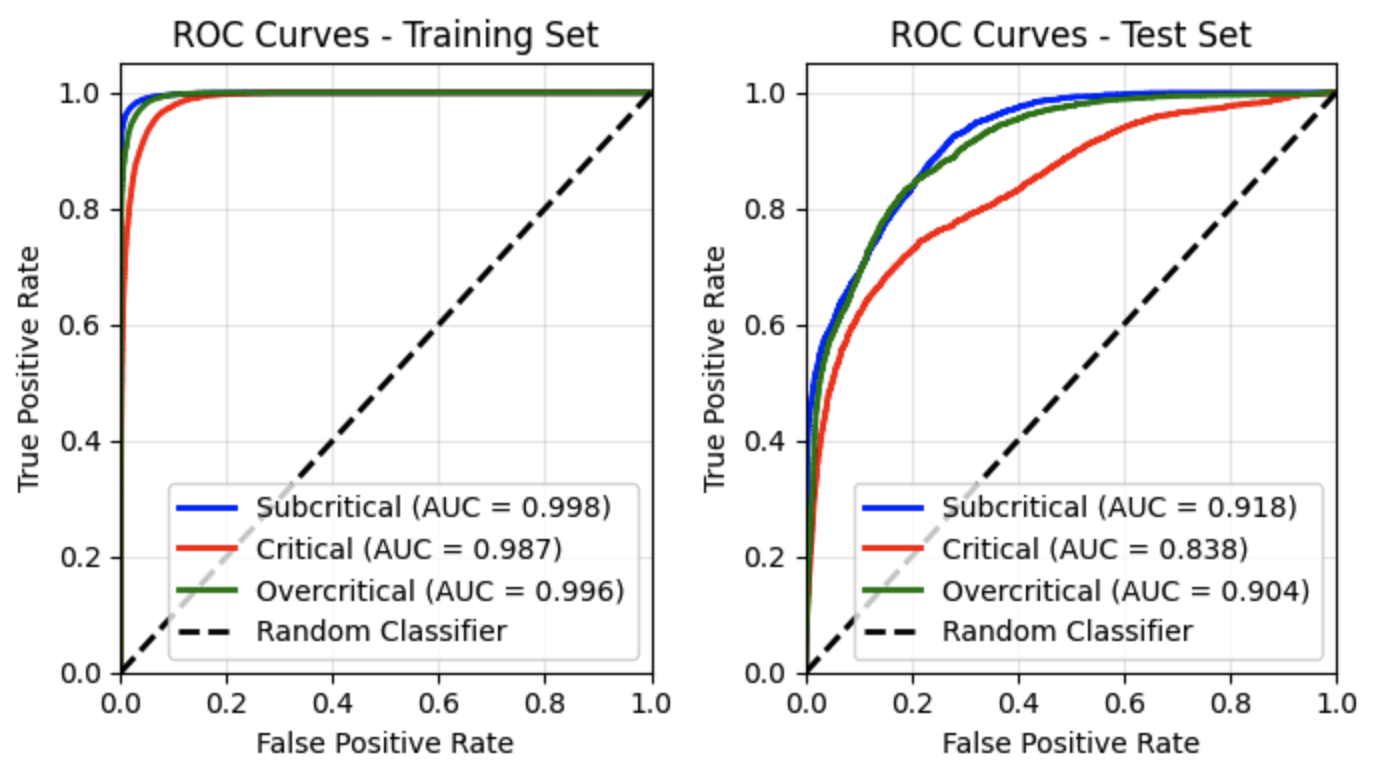}}
\caption{ROC curves for the training and test sets in case B), corresponding to training the discrete thermometer.}
\label{fig_roc}
\end{figure}

To get an idea of the training results and performance of the thermometer, Figure \ref{fig_roc} summarizes the ROC curves for the training set $(80\%)$ and test set (remaining $20\%$) for each of the three classes.  

The true test of performance can be seen in the ROC curves of the test set, which show that the thermometer satisfactorily recognizes subcritical and supercritical samples (AUC of $0.918$ and $0.996$). However, for samples in criticality, performance is somewhat lower (AUC of $0.838$). This is explained by the fact that subcritical systems are generally characterized by being highly noisy, i.e., each agent has its own opinion independent of its neighbors, while supercritical systems are the opposite, tending to have a single large cluster with the same opinion for almost all agents. This makes it easy for the thermometer to detect these distinctive patterns. On the other hand, since the phase transition for these systems is not abrupt, at the edges of this condition (which we have defined at $p=0.75$ and $p=0.85$) the systems are not easily recognizable in one condition or the other. This explains why the thermometer performs less well in recognizing these types of systems. In the subcritical and supercritical regimes, systems exhibit clear spatial signatures (homogeneous disorder vs. large consensus clusters), whereas critical states feature ambiguous mesoscale clusters \cite{carr17} (see Figure \ref{clusters}).

\subsection{Recovery of systems at criticality}
\begin{figure}[t]
\centerline{\includegraphics[scale=0.53]{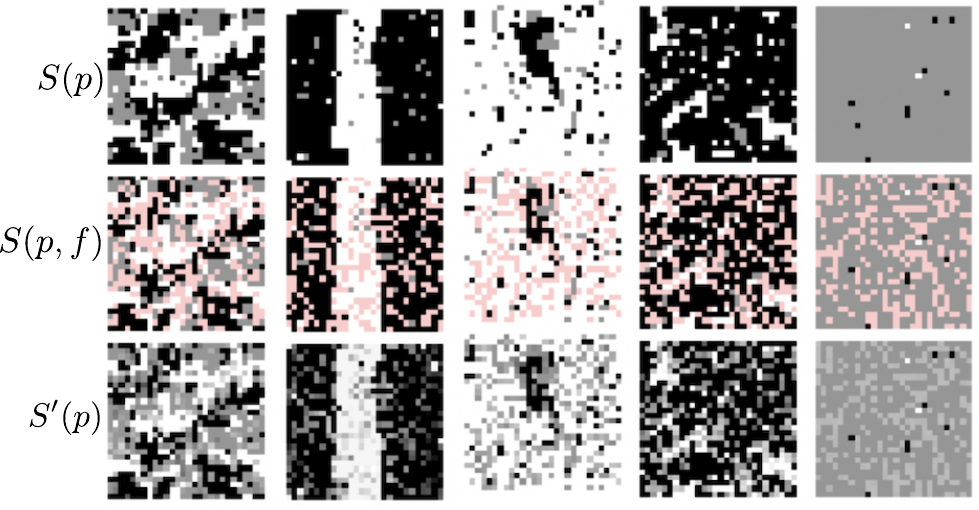}}
\caption{An example showing seven systems (first row $S(p)$), to which a fraction $f=0.3$ of noise equivalent to 235 agents with unknown opinions is added, marked in pink (second row $S(p,f)$), and then the reconstructed system $S'(p)$ in the third row.}
\label{fig_imputation}
\end{figure}

We developed different simulations of MV3 system imputation. Here, we refer to imputation as the process of reconstructing $S(p,f)$ systems in which there is no information about the opinions of a fraction $f$ of the agents. Thus, the challenge for the DBN is to guess those opinions according to the distribution of patterns learned in the previous stage. The process is a mean-field based imputation and it can be summarized as follows: a) System $S(p,f)$ is presented to the DBN. Agents with no information about their opinion are left with a value equal to $-1$. b) The DBN receives the information and transmits it to the last layer, from which the conditional probability $p(v|h_1)$ is calculated as the result. The original opinions of the agents are left clamped in their original opinions and the process is repeated $k$ times. c) $S'(p,f)$ is obtained and compared with the original $S(p)$ to evaluate the error according to the equation (\ref{eq_error}). We focus on samples from systems that are in criticality, which are of greatest interest, so for this analysis we only take samples where $0.75 \leq p \leq 0.85$. Figure \ref{fig_imputation} graphically displays the result obtained from the process.

To gain insight into the behavior of imputation with different architectures, we conducted various experiments with DBN1 $(nH_1=2704, nH_2=144, nH_3=64)$, DBN5 $(nH_1=784, nH_2=255, nH_3=144)$, and DBN7 $(nH_1=256, nH_2=144, nH_3=64)$. When evaluating imputation, it is worth considering some maximum error limits. Remember that $L=N^2$ is the total number of agents in the system and $f$ is the fraction of agents with unknown opinions. We define $e^s$ as the error of the model incorrectly guessing the $fN^2$ opinions of the agents in the worst case, that is, $|s_i - \hat{s}_i| = 1$ and in such a case $e^s = \sqrt{\frac{1^2 fN^2}{N^2}}=\sqrt{f}$. We also define $e^m$ as the error of the model incorrectly guessing the $fN$ opinions, but only when $|s_i - \hat{s}_i| = 0.5$. In this case, $e^m = \sqrt{\frac{(0.5)^2 fN^2}{N^2}}=\frac{\sqrt{f}}{2}$. Table \ref{tab_maxerror} shows a summary of the worst imputation errors that can be obtained.

\begin{table}[t]
\caption{Maximum imputation error limits for a system with $L=N^2$ agents and a fraction $f$ of agents with unknown opinions.}
\begin{center}

\begin{tabular}{lll}
\hline
$f$    & $e^s$ & $e^m$ \\
$0.01$ & $0.100$ & $0.050$ \\
$0.1$  & $0.316$ & $0.158$ \\
$0.3$  & $0.548$ & $0.274$ \\
$0.5$  & $0.707$ & $0.354$ \\
\hline
\end{tabular}

\label{tab_maxerror}
\end{center}
\end{table}

In the Mean-Field approximation for DBN imputation,  We can control how to combine bottom-up (data-driven) and top-down (prior-driven) information at each layer using:

\begin{equation}
    \mathbf{h}_l = \alpha_l \mathbf{h}^{\text{bottom up}}_{l} + (1-\alpha_l) \mathbf{h}^{\text{top down}}_{l}  
\end{equation}

where $l$ indicates the layer, $l=1$ for the first layer, and $l=2$ for the second layer, $\mathbf{h}_{l}$ is the vector of hidden unit values at layer $l$, and $\alpha_1 + \alpha_2 =1$. In other words, the mixing parameters $\alpha_1$ and $\alpha_2$ controls de information flow. The Bottom-Up Information Flow is data-driven and contains incomplete information, while Top-Down Information Flow is prior-driven based on learned global patterns. Thus, if we increase the value of $\alpha_1$, we place more trust in the signals coming from the data and preserve the fine-grained details of the agents' opinions. The first layer (that of the GBRBM) is very rich in real-world information and is necessary to enable detailed reconstructions. On the other hand, if we increase the value of $\alpha_2$, we place more trust in the signals coming from higher layers of compressed representations, which helps to close the gap between the upper compressed layer and the lower layer with details.


\begin{figure}[t]
\centerline{\includegraphics[scale=0.6]{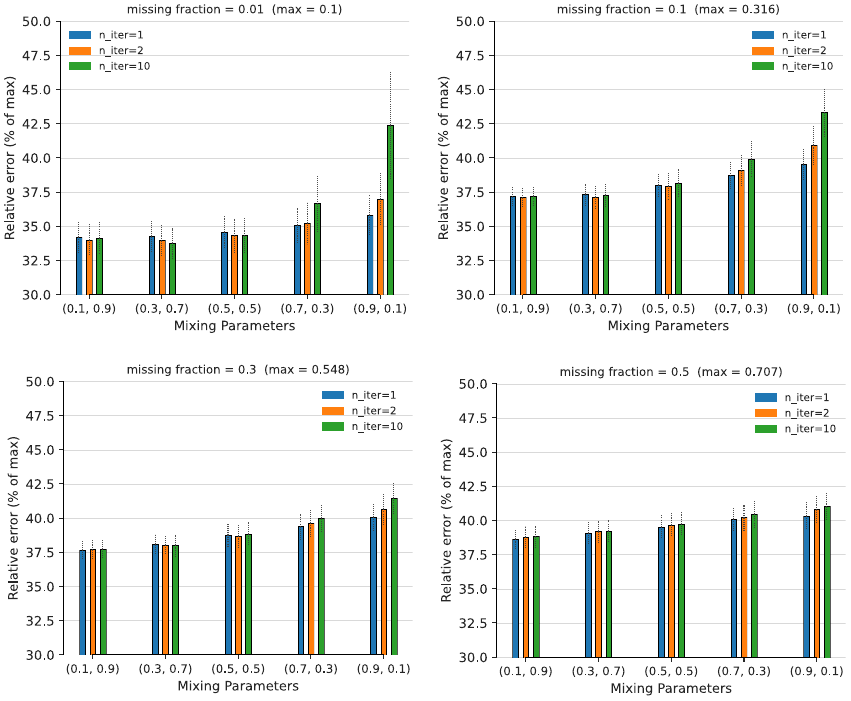}}
\caption{Imputation errors on various $S(p)$ systems using model architecture DBN1 $(nH_1=2704, nH_2=144, nH_3=64)$ for different levels of $f$ and mixing parameters $(\alpha_1, \alpha_2)$. For better comparison, imputation errors are expressed as a percentage of the maximum possible error $e^m$ according to Table \ref{tab_maxerror}.}
\label{fig_imput20704}
\end{figure}

As we can see in Figure \ref{fig_imput20704}, as we increase the level of incomplete information ($f$), the imputation error increases, which is quite expected, reaching values of $40\%$ of the maximum possible error $e^m$. Interestingly, the mixing parameters influence the machine's ability to reconstruct lost agent's opinions. For this expansive-contractive architecture, we see that better results are obtained when we give a greater weight to the information in the initial layer, i.e., a data-driven approach works better. 

\begin{figure}[t]
\centerline{\includegraphics[scale=0.6]{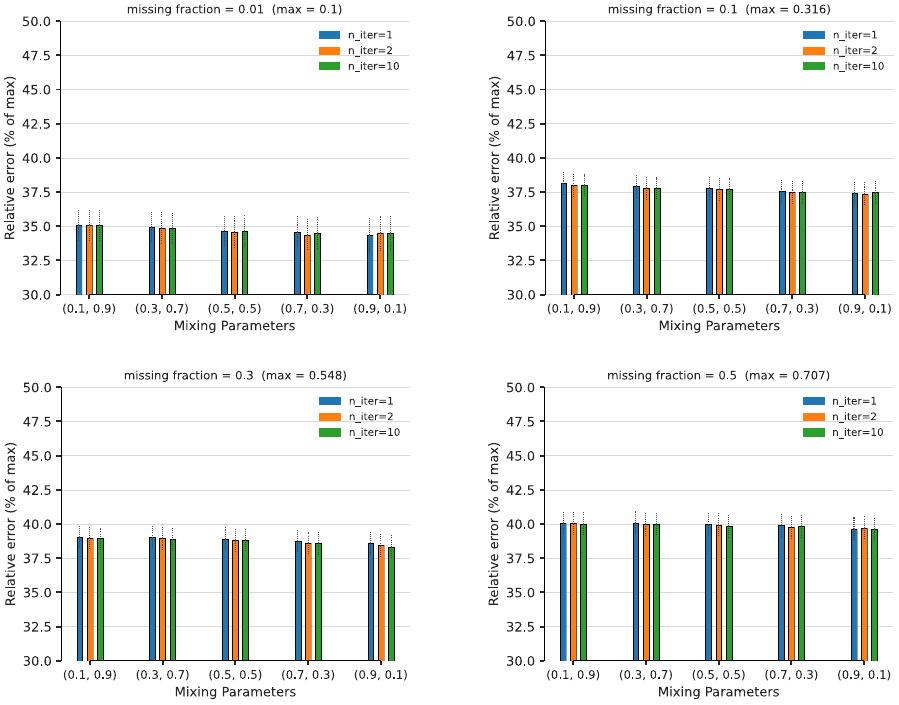}}
\caption{Imputation errors on various $S(p)$ systems using model architecture DBN7 $(nH_1=256, nH_2=144, nH_3=64)$ for different levels of $f$ and mixing parameters $(\alpha_1, \alpha_2)$. For better comparison, imputation errors are expressed as a percentage of the maximum possible error $e^m$ according to Table \ref{tab_maxerror}.}
\label{fig_imput_256_144_64}
\end{figure}

For the DBN7 with compressor architecture (see Figure \ref{fig_imput_256_144_64}), the imputation error behaves differently from the expander-compressor DBN1. We see a very slight decrease in error as we increase the value of $\alpha_1$ and decrease that of $\alpha_2$. Although for the purposes of generating synthetic samples, the expansive architecture in the first layer gives better results, for the compression architecture in all layers it turns out to be competitive when $\alpha_1=0.9$ and $\alpha_2=0.1$.  The intuition behind this result can be understood in that optimal mixing depends on the signal-to-noise ratio (SNR) at each layer. In the case of compression, Layer 3 has high SNR (high compression, essential features), while Layer 1 has lower SNR (many missing values but high dimensionality), thus, a greater weight on the input signals is more convenient.
This contrasts with DBN1, in which by setting values of $\alpha_1=0.2-0.3$ we allow rich top-down information to guide Layer 1, and with $\alpha_2=0.7-0.8$ we maintain the flow of information in the intermediate layers, that is, in the patterns learned for the missing details, maintaining the flow of data through the network.

\subsection{Test of criticality of reconstructed systems}
\begin{figure}[t]
\centerline{\includegraphics[scale=0.6]{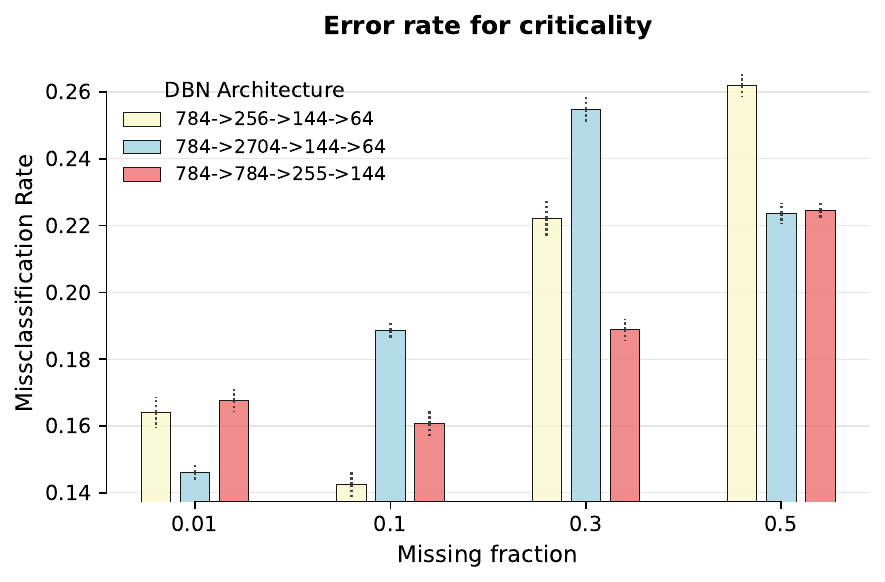}}
\caption{Thermometer misclassification rate for reconstructed systems in criticality $S(p,f)$ at different values of the missing fraction $f$ and three different architectures.}
\label{fig_missclass}
\end{figure}

Using the discrete thermometer described in Section \ref{Section.C}, we evaluate part C (see Figure \ref{fig1}). This analysis should be interpreted with caution, given that we have seen that the thermometer is far from exceptional in its performance, particularly in systems that are precisely at criticality. We adjusted the thermometer's operating point with the aim of maximizing the true positive rate and minimizing the false positive rate (red curve in Figure \ref{fig_roc}) of samples at criticality. The reconstructed samples $S'(p)$ in each of the models were made with a mixing parameter that minimizes the imputation error using $k=1$ Gibbs samplings in the Mean-Field reconstruction. Results are in Figure \ref{fig_missclass}. Mean-Field reconstruction with $k=1$ Gibbs step offers a practical balance. A single forward-backward pass provides sufficient signal from the learned distributions while avoiding the computational cost of multiple iterations.

As we increase the level of lost information, the rate of misclassified critical systems increases. For example, when $f=0.5$, i.e., half of the agents' opinions are lost, the misclassification rate typically increases across all architectures. However, depending on the architecture used, this increase does not appear to be monotonous. For example, DBN1 (expansive-contractive) does not turn out to be the best option, or at least, according to the thermometer, not all reconstructions of this model are recognized in a critical state (with $f=0.3$). The misclassification rate is close to $26\%$, while the contractive DBN7 and DBN5 seem to generate more recognizable samples in a critical state (with $f=0.3$). Thus, on the one hand, previous tests show that expansive architectures tend to have fewer reconstruction errors than contractive architectures. However, when generating criticality samples, they are not necessarily recognized as such. Again, this result warrants further investigation with a larger number of simulations and a more sensitive thermometer to detect the wide variety of systems close to the critical point. It is important to note that the thermometer's reduced sensitivity at the critical point introduces uncertainty in the misclassification rates reported in Figure \ref{fig_missclass}, particularly for higher $f$ values. These results should be interpreted as upper bounds on maintained criticality rather than precise measurements.

\section{Conclusion}
In this work, we analyzed the generative capabilities of DBNs, using a GBRBM as the first input layer, followed by two additional traditional RBM layers for opinion systems governed by majority rule, in which agents can have up to three possible opinion states. We tested different DBN architectures, keeping the number of visible units constant ($784$) and varying the number of hidden units in the GBRBM and in the traditional BBRBMs.
Since we are dealing with opinion systems subject to phase transitions, the samples do not usually present a marked structure between one type of system and another, particularly when the system is in a transition condition in which it is moving from a subcritical to a critical state, or from a critical to a supercritical state. It is precisely at these diffuse edges where a transition change begins to manifest that the patterns are unclear. Proof of this is the somewhat inferior performance of the thermometer in detecting systems in a critical state, compared to its performance in detecting systems in subcritical or supercritical states.

It is also worth mentioning that increasing the number of Gibbs steps in the sample generation process does not improve the DBN's reconstruction capacity (not shown in this study), which may be related to a ``memorization'' effect, in which each stacked RBM memorizes the number of MCMC steps used during learning to estimate the gradient \cite{decelle21}. A dedicated future study, running several experiments with different numbers of Gibbs steps for this particular class of systems with distinct state transitions, could help assess the recovery of the original distributions.

The results indicate that for the purposes of generating or simulating new synthetic systems, expansive DBNs, i.e., those with an increase in the original dimensionality of the problem $(nH_1 > 784)$, tend to offer a lower reconstruction error than contractive architectures, i.e., those in which we decrease the dimensionality of the problem $(nH_1<784)$. On the other hand, when it comes to imputation activities, i.e., when we use DBN to reconstruct systems with unknown or lost opinions, we again see that the expansive architecture tends to have a lower error rate, but is subject to the way we let the information flow through the mixing parameters in the Mean-Field sampling process. With the mixing parameters, we can give greater or lesser weight to information from the input and intermediate layers, achieving a balance between the generic pattern signals learned from the upper layers and the fine-grained data from the input layer. We have seen that these parameters can affect reconstruction errors, especially in expansive architectures.

As a final comment, a system trained on opinion dynamics could provide more robust estimates by learning the underlying structure of how opinions cluster and evolve, rather than simply extrapolating from small samples. Perhaps most intriguingly, the synthetic sample generation capability enables "what-if" scenario exploration \cite{valle}.

In future work, other variants of contrastive divergence with sampling schemes that are more robust to multimodal distributions or regularization applied in the likelihood function could be adapted to establish constraints that meet physical criteria, such as system energy, which could improve the DBN's ability to reconstruct systems of this type that are more consistent with the physics of the system, as well as evaluating the methodology with real data applications in which there is no prior information on the "criticality" condition of the system. Recent generative models such as normalizing flows \cite{kobyzev20, rezende15} and score-based diffusion models \cite{song19, ho20} have shown promise in modeling complex distributions in statistical physics; however, these architectures require either explicit likelihood computation or iterative sampling at inference time. RBMs remain advantageous for constrained generation—the ability to clamp visible units and sample conditionally, as required for imputation—without retraining. Future work may explore hybrid approaches combining DBN inference with score-based refinement.

\bibliographystyle{IEEEtran}
\bibliography{myref}


\end{document}